\documentclass[10pt,twocolumn,letterpaper]{article}

\usepackage[pagenumbers]{cvpr} 

\usepackage{cuted}     
\usepackage{caption}   

\usepackage{booktabs}
\usepackage{amsmath}
\usepackage{xcolor}
\usepackage{colortbl}
\usepackage{adjustbox}
\definecolor{Gray0}{gray}{0.95}
\definecolor{Gray}{gray}{0.9}
\definecolor{Cyan}{rgb}{0.88,1,1}
\usepackage{wrapfig}
\usepackage{array}

\newcommand{\PreserveBackslash}[1]{\let\temp=\\#1\let\\=\temp}
\newcolumntype{C}[1]{>{\PreserveBackslash\centering}p{#1}}
\newcolumntype{L}[1]{>{\PreserveBackslash\raggedright}p{#1}}

\usepackage{multirow}

\definecolor{cvprblue}{rgb}{0.21,0.49,0.74}
\usepackage[pagebackref,breaklinks,colorlinks,allcolors=cvprblue]{hyperref}

\def\paperID{25111} 
\def\confName{CVPR}
\def\confYear{2026}

\title{The Missing Point in Vision Transformers for Universal Image Segmentation}

\author{Sajjad Shahabodini \\
\and
Mobina Mansoori \\
\and
Farnoush Bayatmakou \\
\and
Jamshid Abouei \\
\and
Konstantinos N. Plataniotis  \\
\and
Arash Mohammadi  \\
}

\begin{document}
\maketitle

\begin{strip}
  \centering
  \vspace{-40pt} 
  \includegraphics[width=\textwidth]{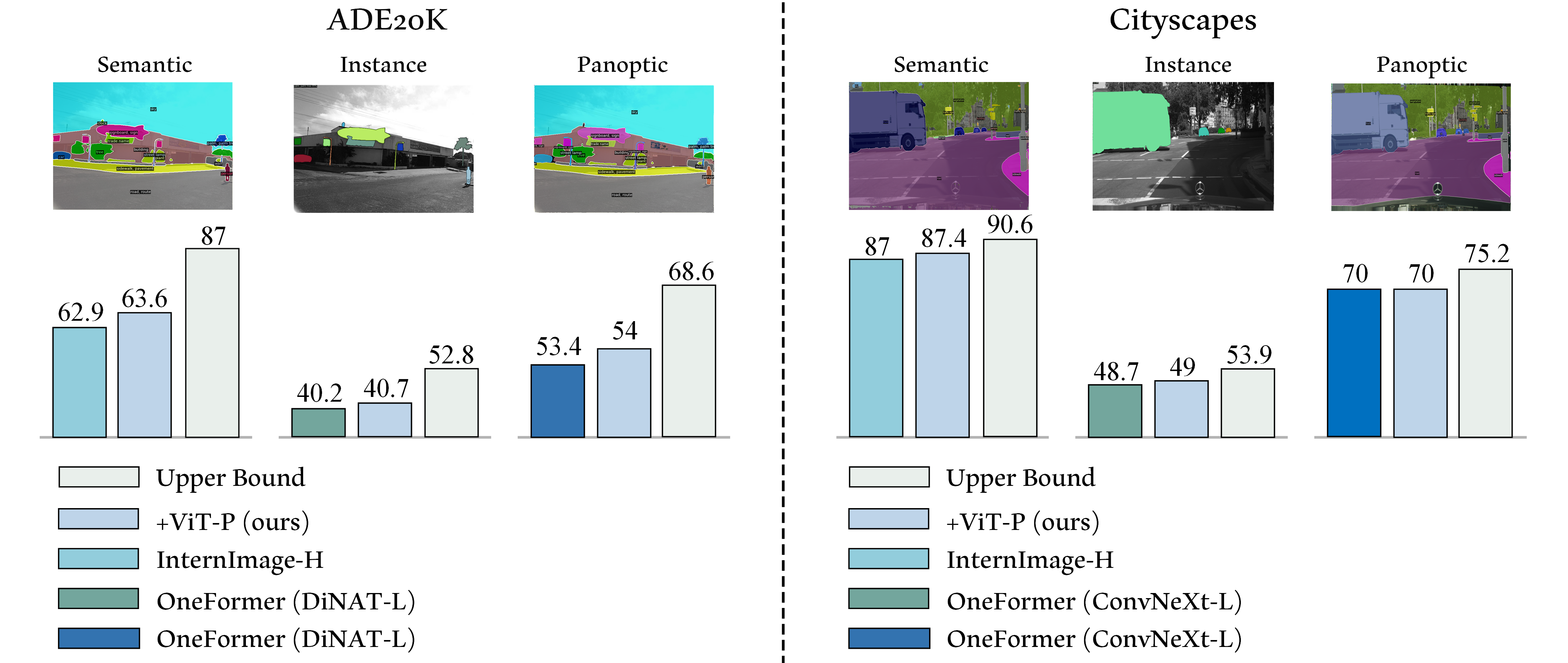} 
  \vspace{-10pt} 
  \captionof{figure}{\textbf{Impact of mask classification on segmentation tasks.} Our ViT-P model improves mask classification accuracy across all three segmentation tasks, leading to enhanced overall performance on the ADE20K~\cite{zhou2017scene}, and Cityscapes~\cite{cordts2016cityscapes} datasets. The upper bound results show the performance when generated masks are classified using ground-truth labels.}
  \label{Figure:figure1}
\end{strip}

\begin{abstract}
Image segmentation remains a challenging task in computer vision, demanding robust mask generation and precise classification. Recent mask-based approaches yield high-quality masks by capturing global context. However, accurately classifying these masks, especially in the presence of ambiguous boundaries and imbalanced class distributions, remains an open challenge. In this work, we introduce ViT-P, a novel two-stage segmentation framework that decouples mask generation from classification. The first stage employs a proposal generator to produce class-agnostic mask proposals, while the second stage utilizes a point-based classification model built on the Vision Transformer (ViT) to refine predictions. ViT-P serves as a pre-training-free adapter, allowing the integration of various pre-trained vision transformers without modifying their architecture, ensuring adaptability to dense prediction tasks. Furthermore, we demonstrate that coarse and bounding box annotations can effectively enhance classification without requiring additional training on fine annotation datasets, reducing annotation costs while maintaining strong performance. Extensive experiments across COCO, ADE20K, and Cityscapes datasets validate the effectiveness of ViT-P, achieving state-of-the-art results with 54.0 PQ on ADE20K panoptic segmentation, 87.4 mIoU on Cityscapes semantic segmentation, and 63.6 mIoU on ADE20K semantic segmentation. The code and pretrained models are available at \href{https://github.com/sajjad-sh33/ViT-P}{https://github.com/sajjad-sh33/ViT-P}.
\end{abstract}    
\section{Introduction}
Image segmentation involves dividing an image into multiple segments or regions, each representing different objects or parts of objects. This segmentation can be based on semantics, such as distinguishing between buildings, vegetation, and the sky, or on specific instances, such as identifying individual cars, pedestrians, or bicycles within an image. Traditional methods~\cite{chen2014semantic,chen2017deeplab,long2015fully,huang2019ccnet} often rely on pixel-wise classification, which struggles to capture the complex relationships between different regions, leading to inaccuracies in object recognition. Recent advancements have led to the development of mask-based segmentation models~\cite{cheng2021per,cheng2022masked,jain2023oneformer}, which significantly improve upon traditional approaches by incorporating global context and spatial dependencies. Instead of classifying each pixel independently, mask-based methods predict a set of binary masks, each corresponding to a single class. This allows for more accurate object separation while preserving spatial coherence. 

Mask-based segmentation models, such as Mask2Former~\cite{cheng2022masked} and InternImage~\cite{wang2023internimage}, generate high-quality masks for object instances, enabling more precise segmentation. However, these models often struggle with classifying the generated masks, particularly in scenarios where object boundaries are ambiguous or class distributions are imbalanced. For example, InternImage~\cite{wang2023internimage} achieves a mean Intersection-over-Union (mIoU) of 62.9\% for semantic segmentation on the ADE20K dataset, but when using ground-truth masks for perfect classification, this improves to 87\%. This gap highlights that while state-of-the-art models excel in mask generation, their classification capabilities remain limited across semantic, instance, and panoptic segmentation tasks.

To address these challenges, we introduce a novel two-stage segmentation approach that divides mask generation and classification, enabling more precise segmentation. The first stage employs a proposal generator to produce class-agnostic mask proposals, ensuring robust mask generation. The second stage, \textbf{ViT-P}, utilizes a point-based classification approach, where classification is performed based on the highest value point of each mask. Built on Vision Transformer (ViT)~\cite{dosovitskiy2020image}, ViT-P incorporates additional point embeddings into the attention mechanism, allowing simultaneous classification of multiple masks in a computationally efficient manner. ViT-P follows a unified training framework, enabling efficient learning across semantic, instance, and panoptic segmentation tasks without requiring task-specific retraining.

In addition to our decoupled framework, we show that the annotation strategy itself can be optimized for segmentation tasks. While fine-grained annotations are crucial for training robust mask generation modules, they are both time-intensive and expensive to obtain \cite{bearman2016s}. In contrast, coarse and bounding box annotations can be generated much more efficiently and still provide valuable supervisory signals for mask classification. In our experiments, augmenting the classification stage with coarse and box annotations not only increased classification performance but also allowed us to minimize the overall annotation cost by enabling model training without requiring additional training or pre-training on segmentation datasets. This dual-annotation approach enables us to use a limited number of fine annotations for high-quality mask proposal generation, while taking advantage of abundant, low-cost coarse data to boost classification accuracy.

We evaluated ViT-P across three fundamental image segmentation tasks: panoptic, instance, and semantic segmentation, using three widely adopted benchmarks: COCO~\cite{lin2014microsoft}, Cityscapes~\cite{cordts2016cityscapes}, and ADE20K~\cite{zhou2017scene}. Our proposed architecture demonstrates competitive or superior performance compared to previous state-of-the-art models, highlighting its effectiveness in enhancing mask classification accuracy across diverse segmentation challenges. Specifically, ViT-P improves semantic segmentation results by +0.7\% on COCO~\cite{lin2014microsoft} and 1.3\% on ADE20K~\cite{zhou2017scene} while utilizing OneFormer~\cite{jain2023oneformer} for mask generation. Additionally, when paired with InternImage~\cite{wang2023internimage}, ViT-P boosts Cityscapes~\cite{cordts2016cityscapes} semantic segmentation accuracy by +0.4\% mIoU. To summarize, our main contributions are:
\begin{itemize}
    \item Decoupling Mask Generation and Classification. We propose a novel two-stage segmentation approach with ViT-P, a pre-training-free adapter for vision transformer backbones that can be pre-trained with not only images but also multi-modal data. ViT-P efficiently adapts these models to downstream dense prediction tasks without modifying their original architecture.
    \item Optimized Annotation Strategy for Classification. We demonstrate that coarse and box annotations can supplement fine annotations, improving classification accuracy without requiring additional training on fine annotation datasets, optimizing annotation costs while maintaining high-quality results.
    \item State-of-the-Art Performance Across Benchmarks. ViT-P achieves \textbf{54.0~PQ} on ADE20K~\cite{zhou2017scene} panoptic segmentation, \textbf{49.0~AP} on  Cityscapes~\cite{cordts2016cityscapes} instance segmentation, and \textbf{53.7~mIoU} on COCO-Stuff-164K~\cite{caesar2018coco} semantic segmentation, setting new performance benchmarks in segmentation.
\end{itemize}
\section{Related Work}
\subsection{Image Segmentation}
In recent years, image segmentation has experienced significant advancements, with various models addressing semantic, instance, and panoptic segmentation challenges. Initially, segmentation was approached as a pixel classification problem, as seen in earlier works such as~\cite{chen2014semantic,chen2017deeplab,cheng2019spgnet,huang2019ccnet,long2015fully}. However, more recent models like MaskFormer~\cite{cheng2021per} and Mask2Former~\cite{cheng2022masked} redefined the task by treating segmentation as a mask classification problem. These models employed a transformer decoder with object queries~\cite{carion2020end} to generate discrete masks. Progress advanced with models like~\cite{hummer2024strong,wang2023image,cai2022reversible,chen2022vision,fang2023eva,wang2023internimage}, each building upon the foundations established by Mask2Former~\cite{cheng2022masked} and significantly enhancing the field. VLTSeg~\cite{hummer2024strong} introduced a novel vision-language pre-training approach, while EVA~\cite{fang2023eva} leveraged masked image modeling, both of which contributed to a more robust segmentation backbone. ViT-Adapter~\cite{chen2022vision} adapted vision transformers to existing frameworks, and InternImage~\cite{wang2023internimage} set new benchmarks by employing a novel dynamic sparse convolution framework. ViT-CoMer~\cite{xia2403vit} refined long-range feature interactions, improving mask-based segmentation accuracy. These advancements have set new standards in image segmentation, showcasing the potential of mask-based segmentation approaches for improved accuracy and efficiency.

\subsection{Universal Image Segmentation}
Further advancements in image segmentation have led to the development of universal models that unify semantic, instance, and panoptic tasks within a single framework. The aim is to create a model capable of handling diverse tasks without relying on specialized architectures or separate training processes. Recent approaches~\cite{gu2023dataseg,jain2023oneformer,xu2024rap,yan2023universal} focus on improving generalization across multiple tasks or datasets rather than designing separate models. The Mask-DINO~\cite{li2023mask} model, for example, is a unified transformer-based framework for object detection and segmentation, extending DINO~\cite{caron2021emerging} by incorporating a mask prediction branch. OneFormer~\cite{jain2023oneformer} was the first framework to surpass state-of-the-art on all three segmentation tasks with a single universal model, integrating them within a unified framework. FreeSeg~\cite{qin2023freeseg} employed one-shot training to optimize an all-in-one model for multiple segmentation tasks, seamlessly handling them in a single architecture. OpenSeeD~\cite{zhang2023simple} unified visual concepts across segmentation and detection tasks and employed conditioned mask decoding to improve mask generation from bounding boxes. Inspired by recent works such as Mask-DINO~\cite{li2023mask} and OpenSeeD~\cite{zhang2023simple} that exploit box annotations to enhance segmentation, we propose leveraging not only fine annotations but also box and coarse annotations to train segmentation models.

\subsection{Transformers and Pre-Training}
Transformers have revolutionized tasks across multiple domains, including natural language processing, computer vision, and speech recognition. Initially introduced for machine translation~\cite{vaswani2017attention}, the vanilla transformer was later adapted for image classification by ViT~\cite{dosovitskiy2020image}. DeiT~\cite{touvron2021training} is an efficient version of ViT~\cite{dosovitskiy2020image} that significantly reduces the data requirements for training transformers by incorporating a distillation token, enabling the model to learn from a teacher network. BEiT~\cite{bao2021beit} and MAE~\cite{he2022masked} further advanced ViT~\cite{dosovitskiy2020image} for self-supervised learning through masked image modeling, demonstrating the versatility of its architecture. Methods such as MoCov3~\cite{chen2024self}, DINO~\cite{caron2021emerging}, DINOv2~\cite{oquab2023dinov2}, iBOT~\cite{zhou2021ibot}, and BEiTv2~\cite{peng2022beit} have further refined self-supervised learning by leveraging contrastive learning, masked image modeling, and knowledge distillation, enhancing the generalization and robustness of vision transformers across diverse tasks. The advent of CLIP~\cite{radford2021learning} showcased the strength of supervised vision-language pre-training in zero-shot robustness, leading to further developments like EVA-CLIP~\cite{sun2023eva}. To enhance ViT's flexibility for segmentation while leveraging advanced pre-training, we propose a pre-training-free adapter that incorporates image priors without modifying the original architecture.

\begin{figure*}[t]
  \includegraphics[width=\linewidth]{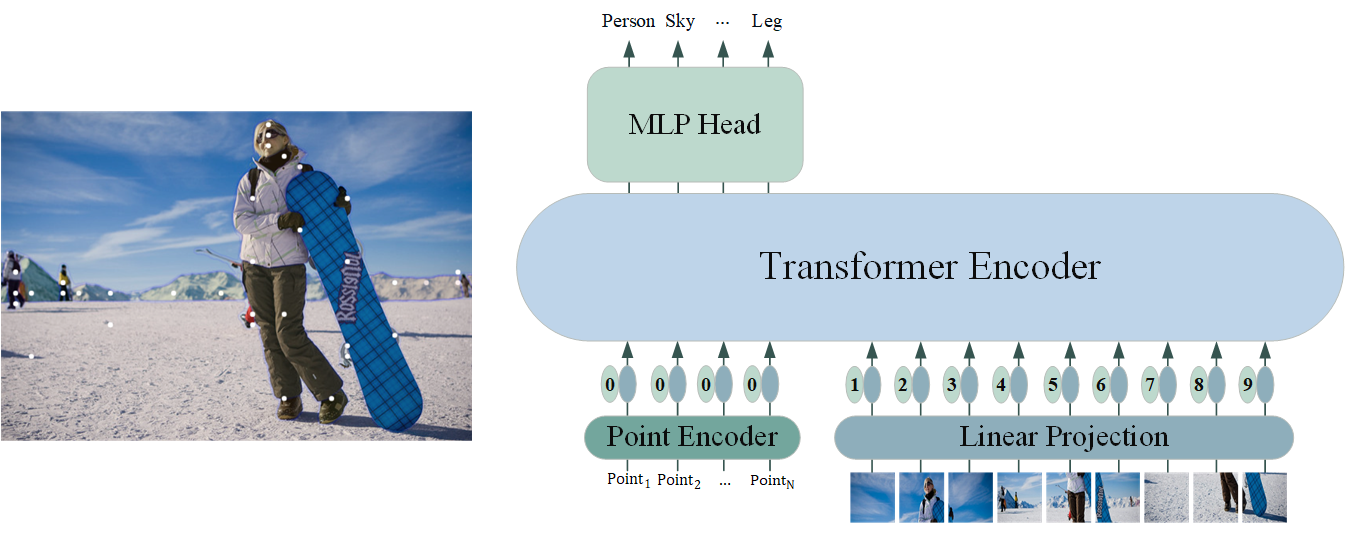}
  \vspace{-25pt} 
  \caption{\textbf{Overview of the ViT-P model architecture.} We propose a two-stage ViT-based approach for universal image segmentation. First, a mask generator produces mask proposals, which ViT-P then classifies by labeling their highest value points using point vision prompting. ViT-P integrates point embeddings with patch embeddings and processes them through a transformer encoder to capture global context, enabling simultaneous per-point classification.}
  \label{fig:Model_Architecture}
  \vspace{-10pt}
  \end{figure*}

\section{Methodology}
In this section, we introduce ViT-P, a universal image mask classification model specifically developed based on vision transformers. Using point positions as an input prompt, ViT-P classifies pre-built masks. We first review a meta-architecture for mask-based classification models and describe how classification is the primary limitation of these segmentation methods. Then, we introduce our improved classifier model, which is key to better segmentation in all semantic, instance, and panoptic tasks. Finally, we describe an efficient inference strategy to benefit from both mask- and point-based classifiers.

\subsection{Mask Segmentation Preliminaries}
MaskFormer \cite{cheng2021per} was the initial model that demonstrated that per-pixel classification is not efficient for image segmentation. This model segments an image by predicting $N$ binary masks, each with a corresponding category label. After the introduction of MaskFormer \cite{cheng2021per}, all further state-of-the-art models such as~\cite{cheng2022masked,wang2023internimage,jain2023oneformer,chen2022vision} followed the same path and tried to enhance the results by improving different parts of the MaskFormer \cite{cheng2021per} model while keeping the same overall structure. 

These models primarily consist of three components: a backbone that extracts low-resolution features, a pixel decoder that up-samples low-resolution backbone features to generate high-resolution per-pixel embeddings, and a transformer decoder that operates on image features to process $L$-dimensional feature vectors. For mask prediction, a Multi-Layer Perceptron (MLP) is typically used to create masks $\left\{m_i \mid m_i \in[0,1]^{H \times W}\right\}_{i=1}^N$ from the per-segment and per-pixel embeddings, where $H \times W$ is the resolution of the image. To obtain class probability predictions $\mathbf{y} \in \mathbb{R}^{N \times (K+1)}$, a linear classifier is applied on top of the feature embeddings, where, $K$ is the number of the category labels. The last element of the probability distribution is an auxiliary “no object” label $\varnothing$ that predicts masks that do not correspond to any of the $K$ categories, thereby removing unnecessary or low-quality masks.

\subsection{From Mask Creation to Classification}
As previously mentioned, all mask-based segmentation models~\cite{cheng2021per,wang2023internimage,li2023mask,fang2023eva,chen2022vision} utilize per-segment embeddings to generate masks and their corresponding classes. Our observations reveal that while these models can produce reasonably accurate masks, their classification performance remains suboptimal, significantly reducing overall segmentation performance. As shown in Figure~\ref{Figure:figure1}, applying perfect classification accuracy for the generated masks (termed the \lq upper bound\rq) would significantly improve overall segmentation performance. However, the results reveal a substantial gap between this upper bound and the actual performance of existing models, primarily due to weak classification capabilities. To bridge this gap, we propose integrating a separate classification model into the mask classification task, thereby enhancing overall segmentation accuracy.

By exploring various possibilities, we discovered that classifying specific points within generated masks can significantly enhance overall segmentation performance (see Section 4.4 for a detailed setup). In particular, we found that the point with the highest value in a generated mask serves as a critical reference. Accurate classification of this point ensures the correct categorization of the entire mask. Notably, this highest-value point is often located near the center of the mask, maintaining a distance from the borders, which makes classification easier and more reliable. The upper bound results in Figure~\ref{Figure:figure1} were also obtained by classifying the generated masks using the label of the highest value point in the ground truth masks. This insight led us to develop a two-stage segmentation model: 1) the first stage employs a proposal generator to produce class-agnostic mask proposals, and 2) then a point-based classification model is employed to classify each mask proposal via classifying its highest value point. The pipeline for the point-based classification model, ViT-P, is described in the following sections.

\subsection{ViT-P Architecture}
Figure \ref{fig:Model_Architecture} provides an overview of the ViT-P model. The overall structure of the ViT-P model is based on the vanilla Vision Transformer~\cite{dosovitskiy2020image}, with additional point embeddings participating in the attention mechanism. While ViT~\cite{dosovitskiy2020image} assigns a category to the input image, the ViT-P model takes $N$ various point positions as inputs, generating category labels for each input point. This design enables the model to handle $N$ mask proposals simultaneously. Similar to ViT~\cite{dosovitskiy2020image}, patch embeddings are used to divide an image $\mathbf{I} \in \mathbb{R}^{H \times W \times C}$ into smaller patches, embedding them as vectors $\mathbf{x}_I \in \mathbb{R}^{N' \times D}$. Here, $C$ represents the number of channels, $N'$ is the number of patches, and $D$ is the dimensionality of the embeddings. Additionally, point embeddings $\mathbf{x}_p \in \mathbb{R}^{N \times D}$ are generated using a point encoder, which consists of a linear layer that transforms input normalized points into vectors with the same dimension as the patch embeddings. The embedding sequence $\mathbf{z}$ is then constructed by concatenating patch embeddings and point embeddings:
\begin{equation}
  \mathbf{z} = \left[ \mathbf{x}_p^1 ; \mathbf{x}_p^2 ; \cdots ; \mathbf{x}_p^N ; \mathbf{x}_I^1 ; \mathbf{x}_I^2 ; \cdots ; \mathbf{x}_I^{N'} \right].
\end{equation}
To preserve positional information, position embeddings are added to the embedding sequence. The same positional embedding is applied to all point embeddings. The embedding sequence is then fed into a standard transformer encoder, which models relationships between embeddings. This enables ViT-P to effectively capture global context and interactions between patches and points. Finally, the generated point feature embeddings are used to compute the predicted point class probabilities $\mathbf{C}_p \in \mathbb{R}^{N \times K}$ with an MLP head.

The ViT-P model leverages pre-trained weights to enhance efficiency and adaptability. While the point embedding layer and MLP head are initialized from scratch, the remaining parameters are derived from a pre-trained transformer, reducing the need for extensive point-labeled data while maintaining robust model performance. Notably, the pre-trained positional embedding of the CLS token is reused for point token positional embeddings.

\subsection{Training Strategy}
\begin{figure*}[t]
  \includegraphics[width=\linewidth]{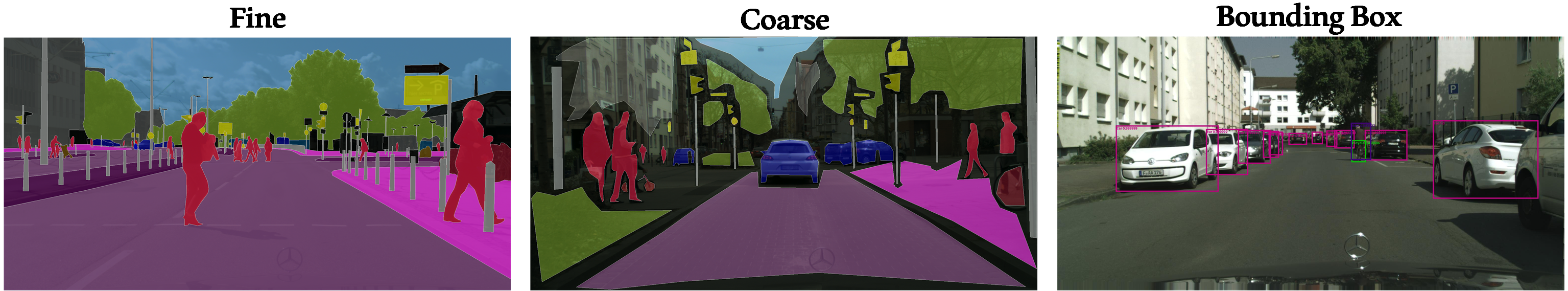}
  \vspace{-10pt} 
  \caption{\textbf{Comparison of annotation types:} Fine segmentation ensures pixel-level precision but is time-intensive, requiring 1 to 2 minutes per object \cite{bearman2016s}. Coarse annotations simplify boundary constraints, offering faster labeling with approximate regions. Box annotations provide the quickest method, typically taking about 10 seconds per object.}
  \label{fig:figure3}
  \vspace{-5pt} 
\end{figure*}

As mentioned earlier, for inference, we utilize masks’ highest value points for better classification results. However, during training, randomly selected points within the image are used to train the ViT-P model. This enables a more robust learning process. Our training methodology leverages three types of annotations (see Figure \ref{fig:figure3}):
\begin{enumerate}
    \item \textbf{Fine Annotations:}  
    Fine annotations use ground truth masks to precisely guide the selection of labeled points. While these annotations can significantly enhance mask classification performance, they require meticulous precision in delineating object boundaries, making their creation extremely time-consuming.
    \item \textbf{Coarse Annotations:}  
    To address the challenges of fine annotations, we employ coarse annotations. These capture approximate object boundaries and can be generated significantly faster, providing a scalable alternative for training.
    \item \textbf{Box Annotations:}  
    Bounding box annotations are utilized exclusively for a pre-training stage. During this phase, we do not sample points. Instead, the model is trained to classify objects using the box coordinates directly as input. A box is represented by its coordinates \mbox{$[x, y, w, h]$} (e.g., center point, width, and height), which are fed into the ViT-P model.
\end{enumerate}
Our training strategy involves first pre-training the ViT-P model using the efficient box annotations. This teaches the model a general understanding of object location and scale. Subsequently, the model is fine-tuned using the more precise point-based inputs from either fine or coarse annotations. To bridge the gap between the pre-training (box inputs) and fine-tuning (point inputs) stages, we adapt the input format. When fine-tuning, a sampled point with coordinates $(x, y)$ is passed to the model as \mbox{$[x, y, 0, 0]$}. This effectively represents the point as a box with zero height and width, maintaining a consistent input structure for the model. In our evaluations, we demonstrate the effectiveness of this combined strategy, pre-training on boxes and fine-tuning on points, in addressing high-quality mask misclassification.

\subsection{Integrating Mask and Point Classifiers}
During inference, we found it beneficial to merge the classification probabilities of the ViT-P model, denoted as $\mathbf{C}_p$, with the class probabilities of the mask generator model, referred to as $\mathbf{C}_m$. The class probabilities $\mathbf{C}_m$ are obtained by removing the "no object" label $\varnothing$ from the class predictions $\mathbf{y}$. Following prior works~\cite{ghiasi2022scaling,xu2023open,yu2023convolutions}, we employ a geometric ensemble to fuse the classification scores:
\begin{equation}
    \mathbf{C}_{\text{fuse}} = \mathbf{C}_m^{(1-\alpha)} \cdot \mathbf{C}_p^{\alpha},
\end{equation}
where $\alpha$ serves as a balance between the predictions of point and mask classifiers. Finally, the "no object" token $\varnothing$ is concatenated to the fused classification scores, ensuring compatibility with both instance and panoptic segmentation tasks.

\begin{table*}[t]
\centering
\footnotesize
\caption{\textbf{Performance comparison on the ADE20K validation set.} ViT-P enhances performance across all three tasks: instance segmentation, panoptic segmentation, and semantic segmentation. FLOPs are calculated using the corresponding crop size. “SS” indicates single-scale testing, while “MS” denotes multi-scale testing. $^\dag$:Models pre-trained on COCO; $^\ddag$: Model is additionally pre-trained on the Objects365 dataset.} \label{tab:ade20k_results}
\adjustbox{width=\linewidth}{
    \begin{tabular}{l l| c c c | c c c c}
        \textbf{Method} & $\underset{\text{(Mask Generator)}}{\textbf{Backbone}}$ & \textbf{\#Params}& \textbf{\#FLOPs}  & $\underset{\text{(Mask Generator)}}{\textbf{Crop Size}}$  & \textbf{PQ}& \textbf{AP} & $\underset{\text{(s.s.)}}{\textbf{mIoU}}$ & $\underset{\text{(m.s.)}}{\textbf{mIoU}}$ \\
        \midrule
        UPerNet \cite{xiao2018unified}  & SwinV2-L \cite{liu2022swin} & — & — & 640$\times$640 & — & — & — & 55.9 \\
        UPerNet \cite{xiao2018unified} & InternImage-L \cite{wang2023internimage} & 256$\mathrm{M}$ & 2526$\mathrm{G}$ & 640$\times$640 & — & — & 53.9 & 54.1 \\
        UPerNet \cite{xiao2018unified} & InternImage-XL \cite{wang2023internimage} & 368$\mathrm{M}$& 3142$\mathrm{G}$ & 640$\times$640 & — & — & 55.0  & 55.3 \\
        \midrule
        MaskFormer \cite{cheng2021per} & Swin-L \cite{liu2021swin} &  212$\mathrm{M}$ &  375$\mathrm{G}$ & 640$\times$640 & — & — & 54.1 & 55.6 \\
        Mask2Former-Panoptic \cite{cheng2022masked} & Swin-L \cite{liu2021swin} & 216$\mathrm{M}$ & 413$\mathrm{G}$ & 640$\times$640 & 48.7 & 34.2 & 54.5 & — \\
        \midrule
        OneFormer \cite{jain2023oneformer} & ConvNeXt-L \cite{liu2022convnet} & 220$\mathrm{M}$ & 389$\mathrm{G}$ & 640$\times$640 & 50.0 & 36.2 & 56.6  & 57.4 \\
        OneFormer \cite{jain2023oneformer} & DiNAT-L \cite{hassani2022dilated} & 223$\mathrm{M}$ & 1768$\mathrm{G}$ & 1280$\times$1280 & 51.5 & 	37.1 & 58.3  & 58.7 \\
        OneFormer$^\dag$ \cite{jain2023oneformer} & DiNAT-L \cite{hassani2022dilated} & 223$\mathrm{M}$ & 1768$\mathrm{G}$ & 1280$\times$1280 & 53.4 & 	40.2 & 58.4  & 58.8 \\
        OneFormer \cite{jain2023oneformer} & ConvNeXt-XL \cite{liu2022convnet} & 372$\mathrm{M}$ & 607$\mathrm{G}$ & 640$\times$640 & 50.1 & 36.3 & 57.4  & 58.8 \\
        \midrule
        OpenSeeD$^\dag$$^\ddag$~\cite{zhang2023simple} & Swin-L \cite{liu2021swin} & 286$\mathrm{M}$ & — & 1280$\times$1280 & 53.7 & 	\textbf{42.6} & 58.4  & — \\
        \midrule
        \rowcolor{Gray0}\textbf{OneFormer+ViT-P} (ours) & DiNAT-L \cite{hassani2022dilated} & 309$\mathrm{M}$ & 1955$\mathrm{G}$ & 1280$\times$1280 & \textbf{51.9} & \textbf{37.8} & \textbf{58.6} & \textbf{59.0} \\
        \rowcolor{Gray}\textbf{OneFormer$^\dag$+ViT-P} (ours) & DiNAT-L \cite{hassani2022dilated} & 309$\mathrm{M}$ & 1955$\mathrm{G}$ & 1280$\times$1280 & \textbf{54.0} & 40.7 & \textbf{59.7} & \textbf{59.9} \\
        \midrule
        \midrule
        UPerNet \cite{xiao2018unified} & SwinV2-G \cite{liu2022swin} & $>$3$\mathrm{B}$ & — & 640$\times$640 & — & — & 59.1 & — \\
        \midrule
        OneFormer \cite{jain2023oneformer} & InternImage-H \cite{wang2023internimage} & 1.1$\mathrm{B}$ &  4193$\mathrm{G}$ & 896$\times$896 & \textbf{54.5} & 40.2 & 60.4  & 60.8 \\
        \midrule
        Mask2Former$^\dag$ \cite{cheng2022masked} & RevCol-H \cite{cai2022reversible} & 2.4$\mathrm{B}$ & — & 640$\times$640 & — & — & 60.4 & 61.0 \\
        Mask2Former$^\dag$~\cite{cheng2022masked}+ViT-Adapter~\cite{chen2022vision} & BEiT-3~\cite{wang2023image} &  1.9$\mathrm{B}$ & — & 896$\times$896 & — & — & 62.0 & 62.8 \\
        Mask2Former$^\dag$ \cite{cheng2022masked} & InternImage-H \cite{wang2023internimage} & 1.31$\mathrm{B}$ &  4635$\mathrm{G}$ & 896$\times$896 & — & — & 62.5  & 62.9 \\
        \midrule
        \rowcolor{Gray0}\textbf{OneFormer+ViT-P} (ours) & InternImage-H \cite{wang2023internimage} & 1.4$\mathrm{B}$ & 4812$\mathrm{G}$ & 896$\times$896 & \textbf{54.5} & \textbf{40.6} & \textbf{61.2} & \textbf{61.6}\\
        \rowcolor{Gray}\textbf{Mask2Former$^\dag$+ViT-P$^\dag$} (ours) & InternImage-H \cite{wang2023internimage} & 1.61$\mathrm{B}$ & 5254$\mathrm{G}$ & 896$\times$896 & — & — & \textbf{63.1} & \textbf{63.6} \\
        \midrule
    \end{tabular}
    }
\vspace{-10pt}
\end{table*}

\begin{table*}[t]
\centering
\footnotesize
\caption{\textbf{Performance comparison on the Cityscapes validation set.} ViT-P surpasses previous state-of-the-art methods in all three instance panoptic and semantic tasks. $^\dag$: Models pre-trained on the Mapillary Vistas dataset; $^\ddag$: Model is additionally pre-trained on the COCO and Objects365 datasets.} \label{tab:Cityscapes_results}
\adjustbox{width=\linewidth}{
    \begin{tabular}{l l c| c c c | c c c c }
        \textbf{Method} & $\underset{\text{(Mask Generator)}}{\textbf{Backbone}}$ & \textbf{Coarse} & \textbf{\#Params} & \textbf{\#Queries}  & $\underset{\text{(Mask Generator)}}{\textbf{Crop Size}}$  & \textbf{PQ}& \textbf{AP} & $\underset{\text{(s.s.)}}{\textbf{mIoU}}$ & $\underset{\text{(m.s.)}}{\textbf{mIoU}}$ \\
        \midrule
        Axial-DeepLab-XL~\cite{wang2020axial} & Axial ResNet-XL~\cite{wang2020axial} & $\times$ & 173$\mathrm{M}$ &  — & 1025$\times$2049 & 64.4 & 36.7 & 80.6 & 81.1 \\
        Panoptic-DeepLab~\cite{cheng2020panoptic}  & SWideRNet~\cite{chen2020scaling} & $\times$&536$\mathrm{M}$ & — & 1025$\times$2049 & 66.4 & 40.1 & 82.2 & 82.9 \\
        \midrule
        Mask2Former-Panoptic \cite{cheng2022masked} & Swin-L \cite{liu2021swin} & $\times$&216$\mathrm{M}$ &  200 & 512$\times$1024 & 66.6 & 43.6 & 82.9 & — \\
        Mask2Former-Semantic \cite{cheng2022masked} & Swin-L \cite{liu2021swin} & $\times$&215$\mathrm{M}$ &  100 & 512$\times$1024 &  — &  — & 83.3 & 84.3 \\
        \midrule
        OneFormer \cite{jain2023oneformer} & ConvNeXt-L \cite{liu2022convnet} & $\times$&220$\mathrm{M}$ &  250 & 512$\times$1024 & 68.5 & 46.5	 & 83.0  & 84.0 \\
        OneFormer$^\dag$ \cite{jain2023oneformer} & ConvNeXt-L \cite{liu2022convnet} & $\times$&220$\mathrm{M}$ &  250 & 512$\times$1024 & \textbf{70.0} & 48.7	 & 84.6  & 85.2 \\
        OneFormer$^\dag$ \cite{jain2023oneformer} & ConvNeXt-XL \cite{liu2022convnet} & $\times$&372$\mathrm{M}$ &  250 & 512$\times$1024 & 69.7 & 48.9 & 84.5  & 85.8 \\
        \midrule
        OpenSeeD$^\ddag$ \cite{zhang2023simple} & Swin-L \cite{liu2021swin} & $\times$&286$\mathrm{M}$ &  300 & 512$\times$1024 & 68.9 & 48.5 & 84.5  & — \\
        \midrule
        \rowcolor{Gray}\textbf{OneFormer$^\dag$+ViT-P} & ConvNeXt-L \cite{liu2022convnet} & \checkmark & 306$\mathrm{M}$ &  250 & 512$\times$1024 & \textbf{70.0} & \textbf{49.0} & \textbf{84.9} & \textbf{85.5} \\
        \midrule
        \midrule
        OneFormer \cite{jain2023oneformer} & InternImage-H \cite{wang2023internimage} & $\times$&1.1$\mathrm{B}$ &  250 & 512$\times$1024 & 70.6 & \textbf{50.6} & 85.1  & 85.7 \\
        \midrule
        Mask2Former~\cite{cheng2022masked}+Depth Anything~\cite{yang2024depth} & ViT-L~\cite{dosovitskiy2020image} & $\times$& — & 200 & 896$\times$896 & — & — & 84.8 & 86.2 \\
        Mask2Former$^\dag$ \cite{cheng2022masked} & InternImage-H \cite{wang2023internimage} & $\times$&1.1$\mathrm{B}$ &  100 & 1024$\times$1024 & — & — & 86.4  & 87.0 \\
        \midrule
        \rowcolor{Gray0}\textbf{OneFormer+ViT-P} & InternImage-H \cite{wang2023internimage} & \checkmark&1.4$\mathrm{B}$ &  250 & 512$\times$1024 & \textbf{70.8} & \textbf{50.6} & \textbf{85.4}  & \textbf{85.9 }\\
        \rowcolor{Gray}\textbf{Mask2Former$^\dag$+ViT-P$^\dag$} & InternImage-H \cite{wang2023internimage} & \checkmark&1.4$\mathrm{B}$ &  100 & 1024$\times$1024 & — & — & \textbf{86.8}  & \textbf{87.4} \\
        \midrule
    \end{tabular}
    }
\vspace{-10pt}
\end{table*}

\begin{table*}[t]
\centering
\footnotesize
\caption{\textbf{Performance comparison on the COCO-Stuff-164K dataset and the COCO val2017.} ViT-P enhances performance across all three tasks: instance segmentation, panoptic segmentation, and semantic segmentation. $^\dag$: Model is additionally trained on the Objects365 dataset.} \label{tab:COCO_results}
\adjustbox{width=\linewidth}{
    \begin{tabular}{l l| c c | c c c | c c }
        & & & & \multicolumn{3}{c|}{\scriptsize{COCO val2017}} & \multicolumn{2}{c}{\scriptsize{COCO-Stuff-164K}} \\
        \textbf{Method} & $\underset{\text{(Mask Generator)}}{\textbf{Backbone}}$ & \textbf{\#Params}& $\underset{\text{(Mask Generator)}}{\textbf{Crop Size}}$ &   \textbf{PQ}  & \textbf{AP} & $\underset{\text{(s.s.)}}{\textbf{mIoU}}$ & $\underset{\text{(s.s.)}}{\textbf{mIoU}}$ & $\underset{\text{(m.s.)}}{\textbf{mIoU}}$ \\
        \midrule
        MaskFormer \cite{cheng2021per} & Swin-L \cite{liu2021swin} &  212$\mathrm{M}$  & 640$\times$640 & 52.7 & — & 64.8 & — & — \\
        Mask2Former-Panoptic \cite{cheng2022masked} &  Swin-L \cite{liu2021swin} & 216$\mathrm{M}$   & 1024$\times$1024 & 57.8 & 48.6 & 67.4 & — & — \\
        \midrule
        kMaX-DeepLab~\cite{yu2022k} & ConvNeXt-L~\cite{liu2022convnet}  & 232$\mathrm{M}$ & 1281$\times$1281 & 58.0 & — & — & — & — \\
        \midrule
        OneFormer \cite{jain2023oneformer} & Swin-L \cite{liu2021swin} & 219$\mathrm{M}$& 1024$\times$1024 & 57.9 &  49.0 & 67.4 & — & — \\
        OneFormer \cite{jain2023oneformer} & DiNAT-L \cite{hassani2022dilated} & 223$\mathrm{M}$& 1024$\times$1024 & 58.0 &  49.2 & 68.1 & — & — \\
        \midrule
        OpenSeeD$^\dag$~\cite{zhang2023simple} & Swin-L \cite{liu2021swin} & 286$\mathrm{M}$  & 1024$\times$1024 & \textbf{59.5} & \textbf{53.2} & 68.6 & — & — \\
        \midrule
        \rowcolor{Gray0}\textbf{OneFormer+ViT-P} (ours) & DiNAT-L \cite{hassani2022dilated} & 309$\mathrm{M}$  & 1024$\times$1024 & 58.0 & 49.5 & 68.6 & — & — \\
        \rowcolor{Gray}\textbf{OneFormer+ViT-P$^\dag$} (ours) & DiNAT-L \cite{hassani2022dilated} & 309$\mathrm{M}$  & 1024$\times$1024 & 58.2 & 49.6 & \textbf{68.8} & — & — \\
        \midrule
        \midrule
        OneFormer \cite{jain2023oneformer} & InternImage-H \cite{wang2023internimage} &  1.35$\mathrm{B}$  & 1024$\times$1024 & 60.0 & 52.0  & 68.8 & — & — \\
        \midrule
        Mask2Former~\cite{cheng2022masked}+ViT-Adapter~\cite{chen2022vision} & BEiTv2~\cite{peng2022beit} &  571$\mathrm{M}$ &  896$\times$896 & —  & —  & — & 52.3 & — \\
        Mask2Former~\cite{cheng2022masked}+ViT-Adapter~\cite{chen2022vision} & EVA~\cite{fang2023eva} &  — &  896$\times$896 & —  & —  & — & 53.4 & — \\
        Mask2Former \cite{cheng2022masked} & InternImage-H \cite{wang2023internimage} & 1.31$\mathrm{B}$ &  896$\times$896 & — & — & — & 52.6  & 52.8 \\
        \midrule
        \rowcolor{Gray}\textbf{Mask2Former+ViT-P} (ours) & InternImage-H \cite{wang2023internimage} & 1.61$\mathrm{B}$ &  896$\times$896 & — & — & — & \textbf{53.5}  & \textbf{53.7} \\
        \midrule
    \end{tabular}
    
    }
\vspace{-10pt}
\end{table*}

\section{Experiments}
We present experiments demonstrating the effectiveness of our universal image segmentation model by benchmarking it against state-of-the-art universal and specialized architectures. Our evaluations include ablation studies across semantic, instance, and panoptic segmentation tasks to validate our design choices. Furthermore, we illustrate the generalizability of our model, achieving outstanding performance on multiple widely used datasets.

\subsection{Datasets and Evaluation Metrics}
\textbf{Datasets.} We evaluated our model using widely recognized image segmentation datasets that cover semantic, instance, and panoptic segmentation tasks. These datasets include \textbf{COCO} \cite{lin2014microsoft}, which comprises 80 "things" and 53 "stuff" categories, \textbf{ADE20K} \cite{zhou2017scene} with 100 "things" and 50 "stuff" categories, and \textbf{Cityscapes} \cite{cordts2016cityscapes} featuring 8 "things" and 11 "stuff" categories. Additionally, \textbf{COCO-Stuff-164K} \cite{caesar2018coco}, with 171 classes, was used for semantic segmentation. Panoptic and semantic segmentation tasks were assessed on the combined "things" and "stuff" categories, while instance segmentation focused exclusively on the "things" categories. Our comprehensive evaluation highlights the robustness of our model across diverse segmentation challenges. 

\textbf{Evaluation Metrics.} To evaluate our model, we employ standard metrics across all three segmentation tasks. For panoptic segmentation, we use the \textbf{PQ} (Panoptic Quality) metric \cite{kirillov2019panoptic}, which evaluates both recognition and segmentation quality. For instance segmentation, the standard \textbf{AP} (Average Precision) metric \cite{lin2014microsoft} is used, while semantic segmentation performance is measured using \textbf{mIoU} (mean Intersection-over-Union) \cite{everingham2015pascal}.

\subsection{Implementation Details}
ViT-P is designed to be compatible with any pre-trained vision transformer backbone. In this study, we choose DINOv2 \cite{oquab2023dinov2} for its strong performance and high-quality results. As a mask proposal generator, we employ OneFormer \cite{jain2023oneformer} as a universal model, while InternImage \cite{wang2023internimage} serves as a non-universal alternative. During training, the mask proposal generator model remains frozen. For large backbone mask generation models, we use DINOv2 Base \cite{oquab2023dinov2}, while for the huge mask generator models, DINOv2 Large \cite{oquab2023dinov2} is used. The number of input points $N$ is set to match the number of mask proposals.

Random points within the input image are selected for classification during training. For inference, the points with the highest confidence scores in the generated masks are used as input to the ViT-P model for mask classification. We train the model using SGD optimizer with a warmup phase and a cosine learning rate scheduler. The training process consists of 30 epochs on COCO \cite{lin2014microsoft} and 60 epochs on ADE20K \cite{zhou2017scene} and Cityscapes \cite{cordts2016cityscapes}, with all setups incorporating 1000 warmup steps followed by a cosine scheduler and a learning rate of $1 \times 10^{-2}$. Additionally, we found it beneficial to apply gradient clipping with a global norm of 1. 

We adopt a crop size of $518\times 518$ for COCO \cite{lin2014microsoft} and ADE20K \cite{zhou2017scene} datasets and $518\times 1036$ for Cityscapes \cite{cordts2016cityscapes}. Training is conducted on four A100 (40GB) GPUs, each utilizing a batch size of 20. Mixed precision floating point is used to increase efficiency. In the inference, we set $\alpha = 0.4$.

\begin{table*}[t]
    \centering
    \footnotesize
    
    \begin{subtable}[t]{0.48\textwidth}
        \centering
        \caption{\small \textbf{Ablation on Number of Input points.} Increasing input points improves training quality.} \label{tab:ablation1}
        \vspace{-2pt}
        \adjustbox{width=\linewidth}{
            \begin{tabular}{L{45pt}|C{45pt}C{45pt}C{45pt}|C{45pt}}
                \textbf{\# of points} & \textbf{PQ} & \textbf{AP} & \textbf{mIoU} & \textbf{\#FLOPs} \\
                \midrule
                $N=50$ & 53.4 & 40.3 & 58.6 & 2243$\mathrm{G}$ \\
                $N=150$ & 53.7 & 40.6 & 59.1 & 1940$\mathrm{G}$ \\
                $N=250$ & 54.0 & 40.7 & 59.7 & 1955$\mathrm{G}$ \\
                $N=350$ & 53.9 & 40.7 & 59.7 & 1969$\mathrm{G}$ \\
            \end{tabular}
        }
    \end{subtable}
    \hfill 
    \begin{subtable}[t]{0.48\textwidth}
        \centering
        \caption{\small \textbf{Ablation on inference mask classification point.} Evaluating point-selection methods for mask classification in ViT-P.} \label{tab:ablation2}
        \vspace{-2pt}
        \adjustbox{width=\linewidth}{
            \begin{tabular}{L{85pt}|C{45pt}C{45pt}C{45pt}}
                & \textbf{PQ} & \textbf{AP} & \textbf{mIoU} \\
                \midrule
                \textbf{highest-value point} & 54.0 & 40.7 & 59.7 \\
                \midrule
                central point & 54.0 & 40.6 & 59.7 \\
                random point & 53.4 & 40.4 & 58.7 \\
            \end{tabular}
        }
    \end{subtable}
    
    \vspace{2pt} 

    \begin{subtable}[t]{0.48\textwidth}
        \centering
        \caption{\small \textbf{Ablation on classification backbone.} evaluating the adaptability of ViT-P with various pre-trained vision transformer architectures.} \label{tab:ablation3}
        \vspace{-2pt}
        \adjustbox{width=\linewidth}{
            \begin{tabular}{L{80pt}|C{50pt}C{50pt}C{50pt}}
                & \textbf{PQ} & \textbf{AP} & \textbf{mIoU} \\
                \midrule
                \textbf{DINOv2 \cite{oquab2023dinov2}} & 54.0 & 40.7 & 59.7 \\
                \midrule
                ViT~\cite{dosovitskiy2020image} & 53.4 & 40.2 & 58.8 \\
                DINOv2+reg \cite{darcet2023vision} & 53.7 & 40.6 & 59.3 \\
                MAE~\cite{he2022masked} & 53.6 & 40.6 & 59.1 \\
            \end{tabular}
        }
    \end{subtable}
    \hfill 
    \begin{subtable}[t]{0.48\textwidth}
        \centering
        \caption{\small Fine, coarse and combined annotation results on the Cityscapes dataset.} \label{tab:ablation0}
        \vspace{-2pt}
        \adjustbox{width=\linewidth}{
            \begin{tabular}{L{80pt}|C{50pt}C{50pt}C{50pt}}
                \textbf{Annotation Type} & \textbf{PQ} & \textbf{AP} & \textbf{mIoU} \\
                \midrule
                \textbf{Fine + Coarse} & \textbf{70.0} & \textbf{49.0} & \textbf{84.9} \\
                \midrule
                Fine & 69.8 & 48.5 & 84.4 \\
                Coarse & 70.0 & 48.8 & 84.8 \\
            \end{tabular}
        }
    \end{subtable}
    
    \vspace{-11pt} 
\end{table*}

\subsection{Main Results}
\textbf{ADE20K.} Without additional pre-training, the ViT-P model exhibits strong adaptability for mask classification on the ADE20K \cite{zhou2017scene} dataset. As shown in Table~\ref{tab:ade20k_results}, our approach is effective across all segmentation tasks, with the greatest impact on semantic segmentation. Using OneFormer \cite{jain2023oneformer} as the mask proposal method, our model achieves improvements, including a +0.6\% increase in PQ, +0.5\% in AP, and +1.3\% in mIoU. Our proposed method reaches 63.6\% mIoU on ADE20K\cite{zhou2017scene}, establishing a new state-of-the-art benchmark.

\textbf{Cityscapes.} The Cityscapes \cite{cordts2016cityscapes} dataset serves as a great benchmark for evaluating the effects of combining fine and coarse annotations, as it provides both annotation types. We conducted experiments assessing ViT-P’s performance under different settings and comparing it with the existing state-of-the-art architectures. Table~\ref{tab:Cityscapes_results} presents results on the Cityscapes \cite{cordts2016cityscapes} validation set, achieving an improvement of +0.2\% in PQ, +0.3\% in AP, and +0.5\% in mIoU metrics. With InternImage~\cite{wang2023internimage} as the mask generation backbone, our model sets a new benchmark, achieving 87.4\% mIoU.

\textbf{COCO.} We evaluate ViT-P against state-of-the-art universal and specialized architectures on the COCO~\cite{lin2014microsoft} val2017  and COCO-Stuff-164K~\cite{caesar2018coco} datasets in Table~\ref{tab:COCO_results}. ViT-P achieves 68.6\% mIoU on COCO~\cite{lin2014microsoft} and 53.5\% on COCO-Stuff-164K~\cite{caesar2018coco}. Additionally, inspired by Mask-DINO~\cite{li2023mask} and OpenSeeD~\cite{zhang2023simple}, which leverage box annotations for improved segmentation, we pre-train ViT-P using the Objects365 \cite{shao2019objects365} dataset. This method enables ViT-P to achieve state-of-the-art performance, reaching 56.8\% mIoU on COCO~\cite{lin2014microsoft}, with slight improvements of +0.2\% for instance segmentation and +0.4\% for panoptic segmentation in OneFormer \cite{jain2023oneformer}.

\subsection{Ablation Studies}
We conducted an in-depth analysis of the ViT-P structure through various ablation studies. Unless otherwise specified, all ablations were performed with the OneFormer \cite{jain2023oneformer} mask generator on the ADE20K \cite{zhou2017scene} dataset.

\textbf{Number of Input Points.} Adjusting the number of input points in the ViT-P model affects both mask proposal classification and overall training performance. Table~\ref{tab:ablation1} shows the results for different input configurations. Insufficient points lead to poor training quality. However, beyond a certain threshold, adding more points does not enhance classification accuracy.

\textbf{Inference Mask Classification Point.} By analyzing the points inside each mask, the ViT-P model classifies mask proposals. In Table~\ref{tab:ablation2}, we evaluate three methods for selecting points within a mask: random selection, the central point (the farthest from mask boundaries), and the highest-value point. Random selection results in a lower mIoU of 58.7\%, while both the central and highest-value points improve performance by 1.0\% mIoU. The highest-value point is preferred as it requires no additional computation and is typically near the mask center.

\textbf{Classification Backbone.} To evaluate the adaptability of the ViT-P model, we conducte an ablation study with four pre-trained vision transformer backbones: a plain ViT~\cite{dosovitskiy2020image}, DinoV2 \cite{oquab2023dinov2}, DinoV2 with additional registers (DinoV2+reg) \cite{darcet2023vision}, and MAE~\cite{he2022masked} as shown in Table~\ref{tab:ablation3}. While the plain ViT~\cite{dosovitskiy2020image} remains effective and enhances mask classification performance, the DinoV2 \cite{oquab2023dinov2} backbone exhibits better compatibility, yielding superior results. Notably, since the vision transformer architecture in the ViT-P model remains unchanged, it can be replaced with more advanced pre-trained transformers in the future, further enhancing its mask classification performance.

\textbf{Coarse Annotations Impact.} Training the ViT-P model on Cityscapes~\cite{cordts2016cityscapes} with combined fine and coarse annotations results in additional accuracy improvements, Table~\ref{tab:ablation0}. Integrating these annotations offers an effective strategy for better segmentation performance.

\section{Conclusion}
In this paper, we proposed ViT-P, a universal image mask classification model based on vision transformers, addressing the limitations of traditional mask-based segmentation methods. By introducing a two-stage approach, ViT-P separates mask generation from classification, using a proposal generator and a point-based classification model. This method enhances the accuracy of classification while maintaining a high-quality mask generation. Our comprehensive experiments demonstrate that ViT-P outperforms existing state-of-the-art segmentation models across multiple benchmarks, achieving new state-of-the-art results on COCO, ADE20K, and Cityscapes datasets. The main advantage of ViT-P is its ability to leverage pre-trained weights while maintaining minimal additional layers, making it efficient and adaptable. This design bridges the gap between standard vision transformers and specialized models for dense prediction tasks. This approach can be further refined to enhance mask classification accuracy, progressively working toward the upper bound, while also improving mask quality to ensure more precise and reliable segmentation.

{
    \small
    \bibliographystyle{ieeenat_fullname}
    \bibliography{main}
}

\clearpage

\section{Upper Bound Analysis}
As defined earlier, the upper bound results for a segmentation model represent the quality of the generated masks, while considering ideal mask classification accuracy. By analyzing these results, we can evaluate the model’s potential to produce precise segmentations regardless of classification errors. In the analysis of the upper bound results, we consider two distinct strategies for assigning labels to the generated masks: 1) leveraging mask proposals to determine the correct label by comparing them with ground truth masks, and 2) identifying the highest-value point within each generated mask and assigning the corresponding ground truth label based on that point. Presented in Table~\ref{tab:appendix1}, the results show that the difference in accuracy between the two strategies is comparatively small. This finding highlights the efficacy of focusing on the most salient point within the mask, suggesting that a segmentation model can achieve robust performance even when label assignment relies on a single high-confidence region rather than the entire mask. Such observations underscore the feasibility of refining segmentation approaches by emphasizing key regions within masks while maintaining overall accuracy.

\renewcommand{\arraystretch}{1.1}
\begin{table*}[!h]
     \centering 
    \caption{\small Comparison of segmentation accuracy between point-based and mask-based label assignment strategies for OneFormer \cite{jain2023oneformer} on three benchmark datasets. The relatively small precision gap indicates that focusing on the most salient region within a mask is nearly as effective as evaluating the entire mask.} \label{tab:appendix1}
    \vspace{+5pt}
    \adjustbox{width=0.9\linewidth}{
    \begin{tabular}{ll|c|ccc}
        \textbf{Model} & \textbf{Backbone} & \textbf{Classification Method} & $\underset{\textbf{mIoU}}{\textbf{COCO}}$ & $\underset{\textbf{mIoU}}{\textbf{Cityscapes}}$ & $\underset{\textbf{mIoU}}{\textbf{ADE20K}}$ \\
        \midrule
        \multirow{2}{*}{OneFormer \cite{jain2023oneformer}} & \multirow{2}{*}{DiNAT-L \cite{hassani2022dilated}} & \text{Point Labeling} & 86.5 & 88.7 & 83.5 \\
        &  & \text{Mask Labeling} & 87.3 & 89.7 & 84.7 \\
    \end{tabular}
    }
\end{table*}

\section{Mask-Prompting vs. Point-Prompting}
While the ViT-P model classifies mask proposals based on input points, an alternative approach is to directly use mask proposals as inputs into a classification model. Prior works, such as MAFT~\cite{jiao2023learning}, explored this approach by feeding the entire mask into a transformer model for labeling. In this section, we compare mask-based prompting with point-based prompting and explain why we favor point-based classification over using the complete mask proposal. To ensure a fair evaluation, both the ViT-P and MAFT \cite{jiao2023learning} models were trained on the ADE20K~\cite{zhou2017scene} dataset using two different backbones, CLIP~\cite{radford2021learning} and DinoV2 \cite{oquab2023dinov2}, with the Mask2Former \cite{cheng2022masked} model serving as the mask generator. 

As shown in Table~\ref{tab:appendix2}, point-based classification, ViT-P, outperforms mask-based prompting, MAFT~\cite{jiao2023learning}, across all three segmentation tasks. By adding only a linear layer to the pre-trained transformer backbone, ViT-P preserves the original architecture. In contrast, the MAFT~\cite{jiao2023learning}  approach integrates $12-L$ additional randomly initialized attention layers into the $L$-layer backbone. This extra layer stack not only increases complexity but also limits the model’s adaptability when paired with standard vision transformer models such as CLIP~\cite{radford2021learning} and DinoV2~\cite{oquab2023dinov2} for dense prediction tasks, ultimately decreasing its performance.

In addition, mask-prompting models such as MAFT~\cite{jiao2023learning} suffer from mask quality during training. Typically, the number of generated masks exceeds the available ground truth masks, leading to the inclusion of additional low-quality mask proposals. Training on these generated masks reduces model accuracy, as overlapping masks introduce noise that negatively impacts learning efficiency. Conversely, training on high-quality ground truth masks creates an inconsistency between the training and inference, as the model is ultimately evaluated on imperfectly generated mask proposals during testing. These inaccuracies introduce inconsistencies that degrade classification performance. In contrast, the ViT-P model benefits from stable input points throughout both training and evaluation, ensuring more consistency. Theses findings highlight the limitations of mask-based prompting relative to point-based classification. Consequently, providing a strong justification for our decision to adopt a point-based approach in this study.

\begin{table*}[!h]
     \centering 
     \footnotesize
    \caption{\small Comparison of point-based, ViT-P, and mask-based, MAFT~\cite{jiao2023learning}, prompting on ADE20K~\cite{zhou2017scene} segmentation using CLIP~\cite{radford2021learning} and DinoV2~\cite{oquab2023dinov2} backbones. } \label{tab:appendix2}
    \vspace{+5pt}
    \adjustbox{width=0.8\linewidth}{
    \begin{tabular}{L{120pt}|c|ccc}
        \textbf{Method} & \textbf{Classification Backbone} & \textbf{PQ} & \textbf{AP} & \textbf{mIoU} \\
        \midrule
        Mask2Former \cite{cheng2022masked} & — & 48.7 & 34.2 & 54.5 \\
        \midrule
        Mask2Former \cite{cheng2022masked}+MAFT \cite{jiao2023learning} & \multirow{2}{*}{CLIP~\cite{radford2021learning}} & 47.6 & 33.3 & 54.6 \\
        Mask2Former \cite{cheng2022masked}+ViT-P &  & \textbf{49.0} & \textbf{34.4} & \textbf{55.3} \\
        \midrule
        Mask2Former \cite{cheng2022masked}+MAFT \cite{jiao2023learning} & \multirow{2}{*}{DinoV2 \cite{oquab2023dinov2}} & 48.6 & 34.2 & 54.8 \\
        Mask2Former \cite{cheng2022masked}+ViT-P &  &\textbf{49.3} & \textbf{ 34.7 }& \textbf{56.1} \\
    \end{tabular}
    }
\end{table*}

\section{Limitations}
While the proposed model demonstrates improved classification accuracy for generating mask proposals, a performance gap persists relative to the upper bound. This gap highlights the limitations of current mask classification strategies and indicates that further work is needed to enhance overall segmentation performance. Additionally, while ViT-P improves semantic segmentation performance, gains in instance and panoptic segmentation are marginal. Mask selection for individual objects is still a challenge for the model, highlighting the need for better instance selection methods.

Another limitation of the ViT-P model is its reliance on ensembling with the mask proposal model's classification probabilities. Since mask generator models are trained on ground truth masks, mask proposal models tend to assign higher probability values to higher-quality masks, a nuance that is not present in point-based classifiers. Consequently, the ViT-P model achieves strong segmentation performance only when combined with the mask proposal’s probabilities. Eliminating this dependency would require more improvements in the point-based classification results, as the reported upper bound results are obtained without such ensembling.

\section{Qualitative Results}
\begin{figure*}[t]
  \centering
  \includegraphics[width=0.8\linewidth]{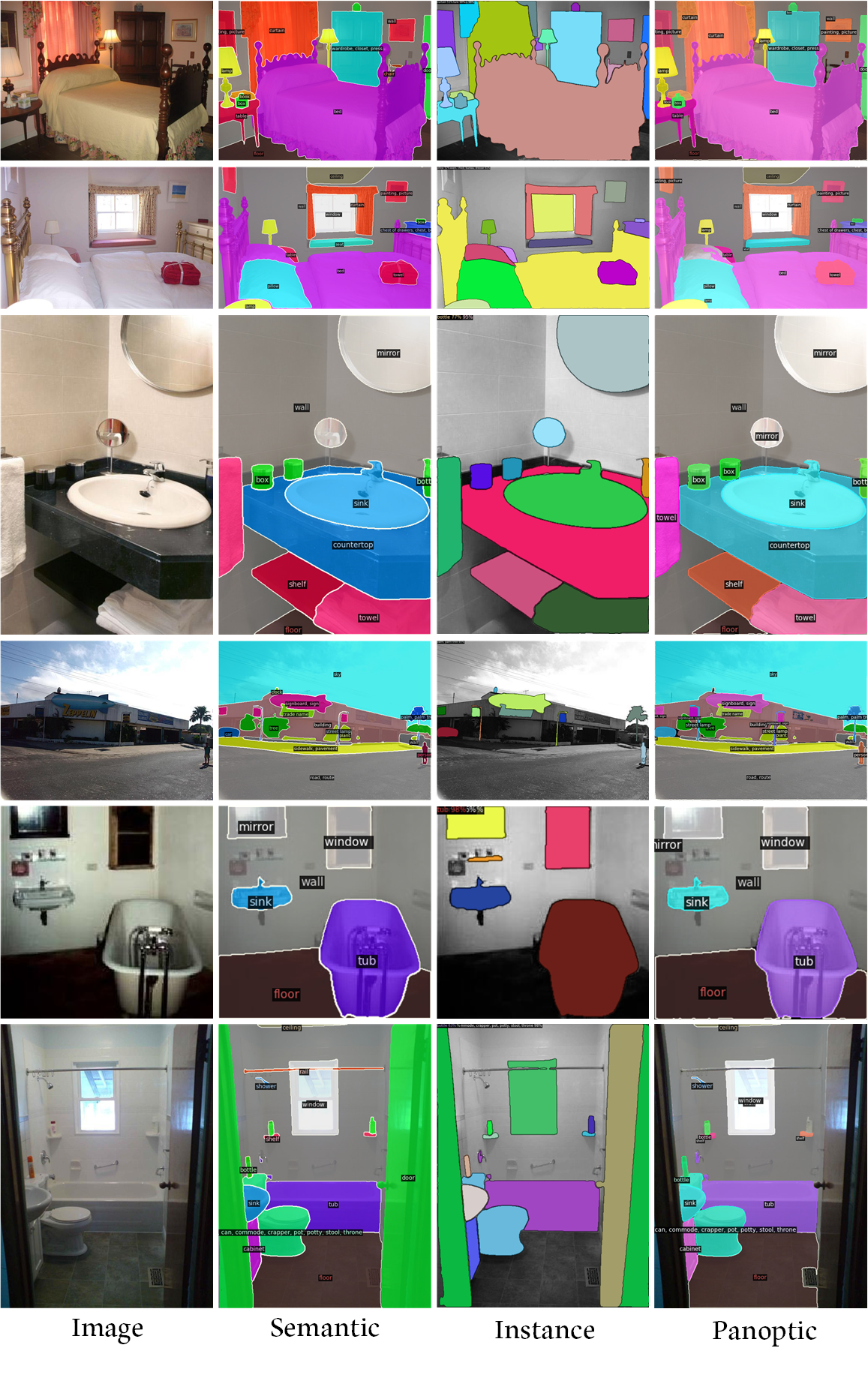}
  \caption{\textbf{Segmentation examples on the ADE20K dataset.}}
  \label{fig:appendix1}
  \end{figure*}

\begin{figure*}[t]
  \centering
  \includegraphics[width=0.75\linewidth]{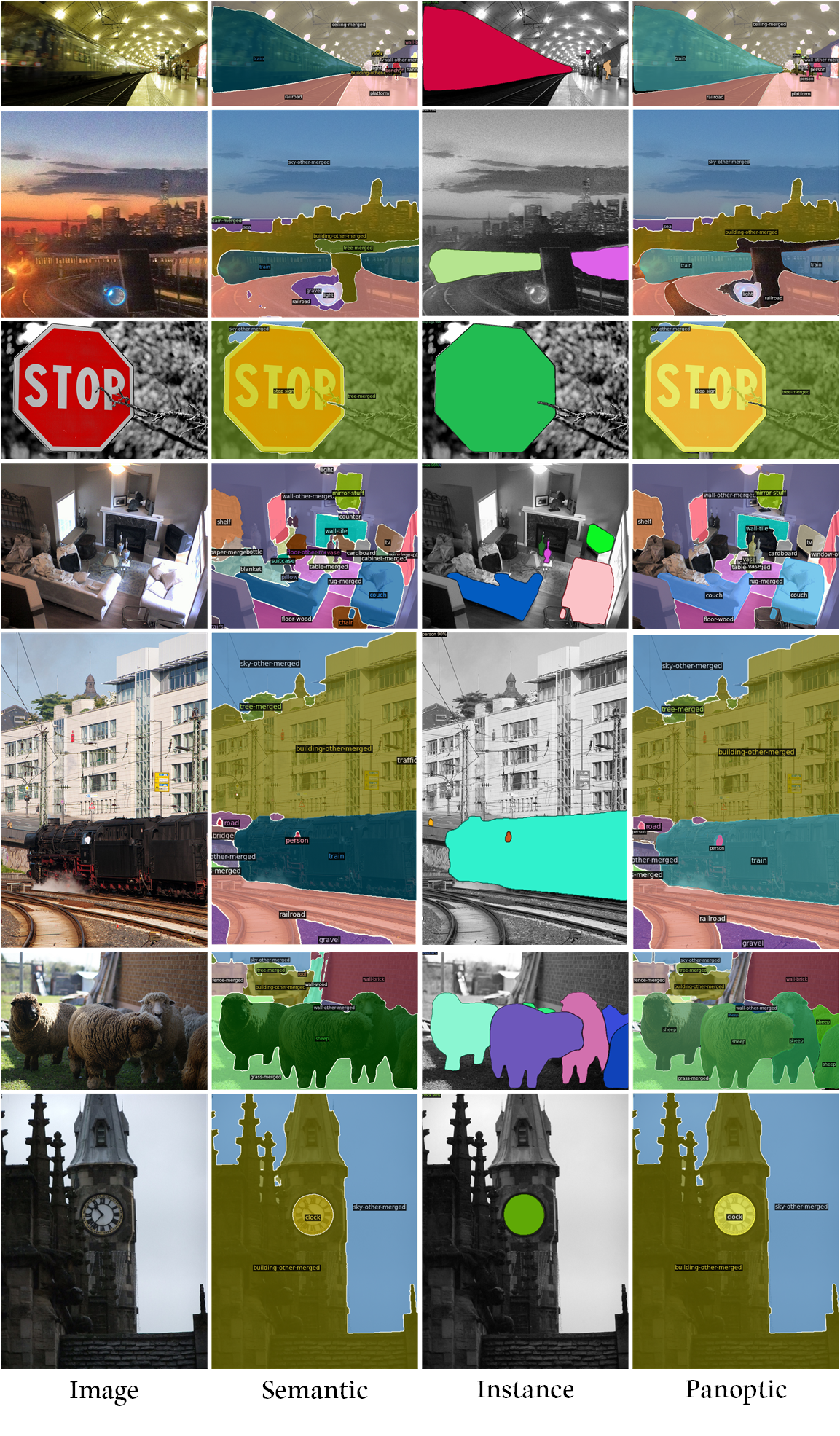}
  \caption{\textbf{Segmentation examples on the COCO dataset.}}
  \label{fig:appendix2}
  \end{figure*}

\begin{figure*}[t]
  \centering
  \includegraphics[width=0.8\linewidth]{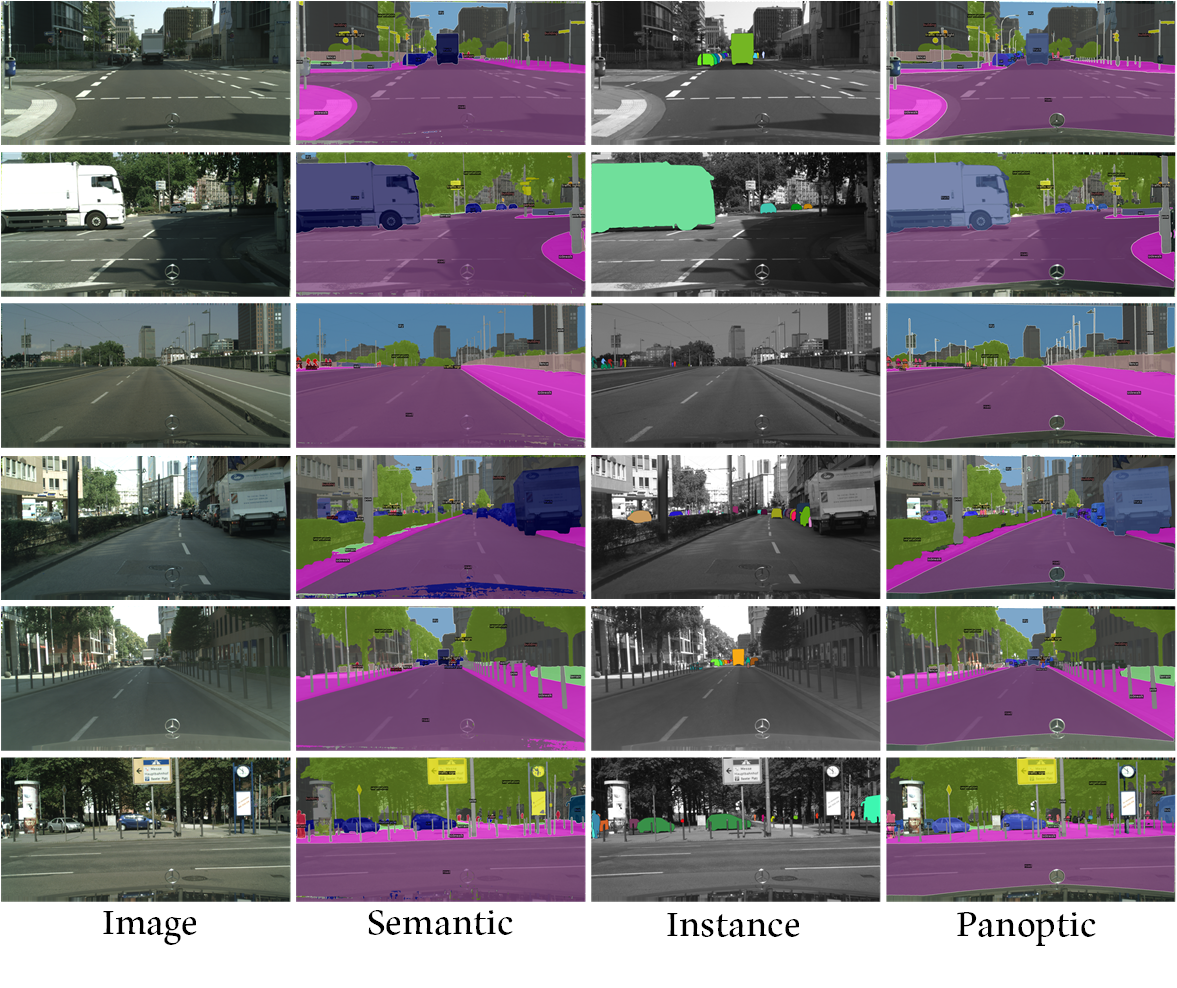}
  \caption{\textbf{Segmentation examples on the Cityscapes dataset.}}
  \label{fig:appendix3}
  \end{figure*}

We present segmentation examples on three benchmark datasets, ADE20K~\cite{zhou2017scene}, COCO~\cite{lin2014microsoft}, and Cityscapes~\cite{cordts2016cityscapes}, illustrating ViT-P model's performance, in  Figure \ref{fig:appendix1}, \ref{fig:appendix2}, \ref{fig:appendix3}. 

\end{document}